\documentclass{article}

\PassOptionsToPackage{numbers, compress}{natbib}
\usepackage[final]{neurips_2024}

\usepackage[utf8]{inputenc} 
\usepackage[T1]{fontenc}    
\usepackage{hyperref}       
\usepackage{url}            
\usepackage{booktabs}       
\usepackage{amsfonts}       
\usepackage{nicefrac}       
\usepackage{microtype}      
\usepackage{xcolor}         
\usepackage{graphicx}
\usepackage{amsmath}
\usepackage{multirow}
\usepackage{multicol}
\usepackage{wrapfig}
\usepackage{tabularx}

\title{StreamingDialogue: Prolonged Dialogue Learning via\\ Long Context Compression with Minimal Losses}

\author{%
Jia-Nan~Li$^{1}$\thanks{The first two authors contributed equally.}
  \quad Quan~Tu$^{1}$\footnotemark[1] \quad Cunli Mao$^{2}$ \quad
\textbf{Zhengtao~Yu}$^{2}$\footnotemark[2] \quad \textbf{Ji-Rong~Wen}$^{1}$ \quad \textbf{Rui~Yan}$^{1}$\thanks{Corresponding authors: Rui~Yan (ruiyan@ruc.edu.cn) and Zhengtao~Yu (ztyu@hotmail.com).} \\
  $^{1}$ Gaoling School of Artificial Intelligence, Renmin University of China\\
  $^{2}$ Kunming University of Science and Technology\\
  \texttt{\{lijianan, quantu, jrwen, ruiyan\}@ruc.edu.cn} \\
\texttt{maocunli@163.com}, \texttt{ztyu@hotmail.com}\\
}

\begin{document}

\maketitle

\begin{abstract}
  Standard Large Language Models (LLMs) struggle with handling dialogues with long contexts due to efficiency and consistency issues. According to our observation, dialogue contexts are highly structured, and the special token of \textit{End-of-Utterance} (EoU) in dialogues has the potential to aggregate information. We refer to the EoU tokens as ``conversational attention sinks'' (conv-attn sinks). Accordingly, we introduce StreamingDialogue, which compresses long dialogue history into conv-attn sinks with minimal losses, and thus reduces computational complexity quadratically with the number of sinks (i.e., the number of utterances). Current LLMs already demonstrate the ability to handle long context window, e.g., a window size of 200K or more. To this end, by compressing utterances into EoUs, our method has the potential to handle more than 200K of utterances, resulting in a prolonged dialogue learning. In order to minimize information losses from reconstruction after compression, we design two learning strategies of short-memory reconstruction (SMR) and long-memory reactivation (LMR). Our method outperforms strong baselines in dialogue tasks and achieves a 4 $\times$ speedup while reducing memory usage by 18 $\times$ compared to dense attention recomputation.\footnote{Code: \url{https://github.com/JinaLeejnl/StreamingDialogue}}
\end{abstract}

\section{Introduction}

\begin{figure*}[!h]
\vspace{-6pt}
  \centering
  \includegraphics[width=0.95\linewidth]{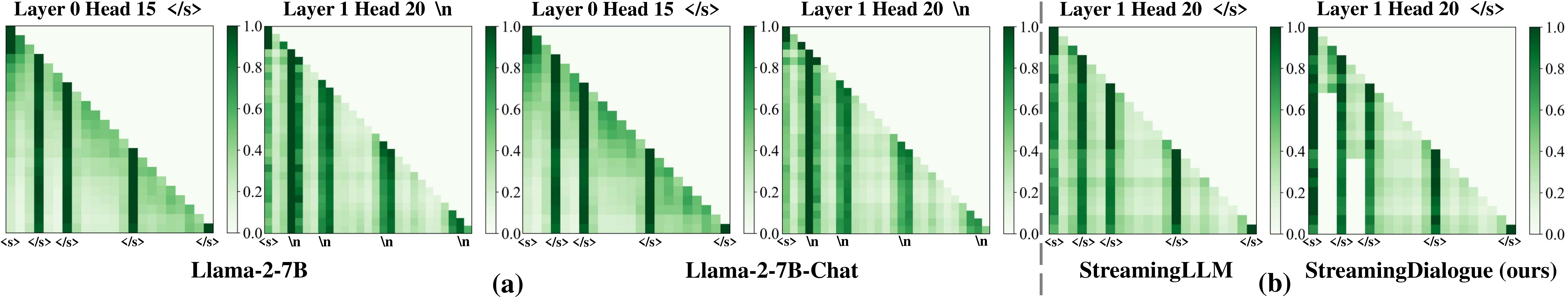}
  \caption{Attention map visualization. (a) Llama-2-7B/Chat with ``</s>'' and ``\textbackslash{}n'' as EoU (``</s>'' counts as one token, ``\textbackslash{}n'' as two).
(b) StreamingLLM versus StreamingDialogue attention on Llama-2-7B with ``</s>'' as EoU.}
  \label{fig:observe}
\end{figure*}


Large Language Models (LLMs) \citep{raffel2020exploring, yang2023baichuan, achiam2023gpt, liu2024skepticism, anil2023palm} are rapidly advancing. However, their performance is constrained by context size during pre-training. For example, with a context size of 4,096, the inference capability of LLaMA2 \citep{touvron2023llama} sharply drops when the context length exceeds the preset limit. Moreover, the attention mechanism \citep{vaswani2017attention} leads to quadratic growth in computational complexity with text length, increasing GPU memory usage and slowing generation speed. As LLMs find widespread use in various conversational applications \citep{yuhan2023unleashing, lee-etal-2021-dialogue, wu2024unify}, these limitations become particularly severe for dialogue tasks \citep{tu-etal-2022-misc, bebensee-lee-2023-span, lv-etal-2023-dialogps}, rendering standard LLMs infeasible for supporting prolonged dialogues with long conversation histories.


In order to support conversations with long contexts, a natural solution is to reduce the computation of inter-token correlations by modifying the implementation of attention. \citet{beltagy2020longformer} proposed local attention, which confines attention within a $k$ window size, reducing computational complexity to linear. However, when the text length exceeds $k$, the generation performance substantially declines. StreamingLLM \citep{xiao2023efficient} enhanced long context streaming by introducing the concept of "attention sinks." This approach builds on local attention, allowing initial tokens to be consistently attended to, which supports stable long-term interactions and efficient generation. However, StreamingLLM continuously updates the cached information within the fixed-size window, excluding initial tokens. These initial tokens are blind to subsequent tokens during the auto-regressive generation process. Consequently, as the dialogue context lengthens, historical information is progressively lost, which is detrimental to dialogue consistency and severely impacts the user experience.



We introduce StreamingDialogue, a method designed for efficient conversations with enhanced long-term memory capabilities. In dialogue contexts, we observe an interesting phenomenon: tokens used to separate utterances (namely End-of-Utterance, EoU), such as ``</s>'' and ``\textbackslash{}n,'' tend to aggregate more attention than other tokens (for details, refer to Figure \ref{fig:observe} (a) and see \S\ref{Generality} for further analysis). We refer to these separator tokens as \textbf{``conversational attention sinks''} (conv-attn sinks). Figure \ref{fig:observe} (b) demonstrates that, in contrast to the highly dispersed attention pattern of StreamingLLM, StreamingDialogue maintains focus on critical positions like conv-attn sinks, thereby utilizing them to aggregate utterance information, compressing lengthy dialogues to only require caching conv-attn sinks' key-values to improve efficiency and reduce memory consumption.

Specifically, in the long-term generation, we preserve conv-attn sinks to memorize historical dialogues for retrieval. Additionally, caching both the first token and the previous and current utterances is crucial to ensure stable output beyond a certain inference length and to facilitate the smooth generation of consecutive replies. Beyond these measures, we introduce two self-learning strategies to better characterize the conv-attn sinks: (1) we devise a reconstruction task, where the reconstruction process can only attend to the conv-attn sink of the target utterance, thereby encouraging the conv-attn sink to restore information from the target sentence, namely \textbf{short-memory reconstruction} (SMR); (2) we propose a recall task, treating the final utterance as a query and attending solely to conv-attn sinks in the dialogue history to retrieve the matching response, thus prompting the model to reactivate information from lengthy dialogues, named as \textbf{long-memory reactivation} (LMR). These two tasks will be jointly optimized before dialogue learning.

Experiments on widely-used dialogue datasets demonstrate that our proposed method outperforms other sparse attention and memory-enhancement methods (in terms of evaluation metrics of Perplexity, BLEU, ROUGE, Distinct, USL-H, and Dial-M). In terms of efficiency, our method achieves a 4 $\times$ speedup and an 18 $\times$ reduction in memory usage compared to dense attention with recomputation. In particular, currently some LLMs support handling long contexts, such as Claude 2.1\footnote{\url{https://www.anthropic.com/news/claude-2-1}} with a 200K context window. In this way, leveraging our method with such long context LLMs enables the completion of numerous utterances within the conversation session, which indicates one big step towards prolonged dialogue learning with long contexts. In summary, our main contributions are as follows:

(1) We discover that EoU tokens have the potential to aggregate utterance information. By defining these separator tokens as ``conv-attn sinks,'' we propose StreamingDialogue, which efficiently handles long context by only caching the first token, conv-attn sinks, and tokens from the most recent two utterances.

(2) We propose two learning strategies: short-memory reconstruction (SMR) and long-memory reactivation (LMR), enhancing the capability of conv-attn sinks to aggregate information and the ability to store historical information.

(3) We demonstrate that StreamingDialogue significantly reduces computational complexity experimentally, ensuring the efficiency of streaming conversations.

\section{Related work}
StreamingDialogue efficiently handles long context, improving the model's long-term memory for conversation history. Existing methods for processing long context in transformer-based models broadly fall into three categories: efficient transformer design, long-term memory enhancement, and length extrapolation techniques.

\subsection{Efficient transformers}
Due to attention's computational bottleneck in transformers, some methods aim to explore efficient attention mechanisms. Solutions include trading accuracy for speed, e.g., Longformer \citep{beltagy2020longformer} employs sliding window attention, expanding the receptive field with a dilated sliding pattern and optionally integrating global attention. BP-Transformer \citep{ye2019bp} balances complexity and capacity with fine-to-coarse attention across multiple scales using binary partitioning. Linformer \citep{wang2020linformer} approximates self-attention with a low-rank matrix, simplifying operations to linear ones. LongLoRA \citep{chen2023longlora} uses block-wise attention and token shifting to enhance communication between blocks. Another solution lies in system-level optimizations, e.g., FlashAttention \citep{dao2022flashattention, dao2023flashattention} optimizes memory access by perceptually reading and writing, improving efficiency without sacrificing accuracy. However, these methods don't preserve dialogue history or expand the context window sufficiently for prolonged dialogue with long-term memory.

\subsection{Long-term memory}
Some methods enhance models' long-term memory to improve long-text modeling. One approach is introducing recurrent mechanisms into attention, enabling the model to maintain information over long sequences. For example, Transformer-XL \citep{dai2019transformer} introduces segment-level recurrence, reusing previous time step hidden states to model long dependencies. $\infty$-former \citep{martins-etal-2022-former} employs continuous-space attention for arbitrary context modeling with fixed computational cost. Another approach is utilizing existing models as interfaces to external knowledge bases, enhancing contextual input and long-term memory through reading and writing to these bases during inference \citep{huang2023advancing}, e.g., MemGPT \citep{packer2023memgpt} employs hierarchical memory for LLMs, optimizing information transfer between context windows and external storage. However, they require retraining LLMs from scratch or additional information retrieval, lacking efficiency.

\subsection{Length extrapolation}
Length extrapolation in models refers to their ability to maintain good performance beyond the training length during inference. A mainstream solution is based on position encoding. LLMs \citep{roziere2023code, bai2023qwen, 2023llama} employ rotary position embedding (RoPE) \citep{su2024roformer, chen2023fortify, liu2023scaling, lin2024mixture} for length extrapolation without fine-tuning. Initially introduced by \citet{chen2023extending}, position interpolation proportionally extends the inference length by reducing rotation angles. NTK-aware\footnote{\url{https://www.reddit.com/r/LocalLLaMA/comments/14lz7j5/ntkaware_scaled_rope_allows_llama_models_to_have/}} and NTK-by-parts\footnote{\url{https://github.com/jquesnelle/yarn/pull/1}} interpolations balance high and low-frequency information to optimize performance.
YaRN \citep{peng2023yarn} combines NTK-by-parts interpolation with an attention distribution correction strategy, reducing rotation angles for low frequencies and adjusting attention distribution. Additionally, randomized position encoding \citep{ruoss-etal-2023-randomized} extends context exposure by decoupling pre-training length from inference length, utilizing random positions during training for broader context coverage. Due to current methods' inability for infinite length extrapolation, they're unsuitable for prolonged dialogue in streaming applications.

\begin{figure*}[!h]
  \centering
  \includegraphics[width=\linewidth]{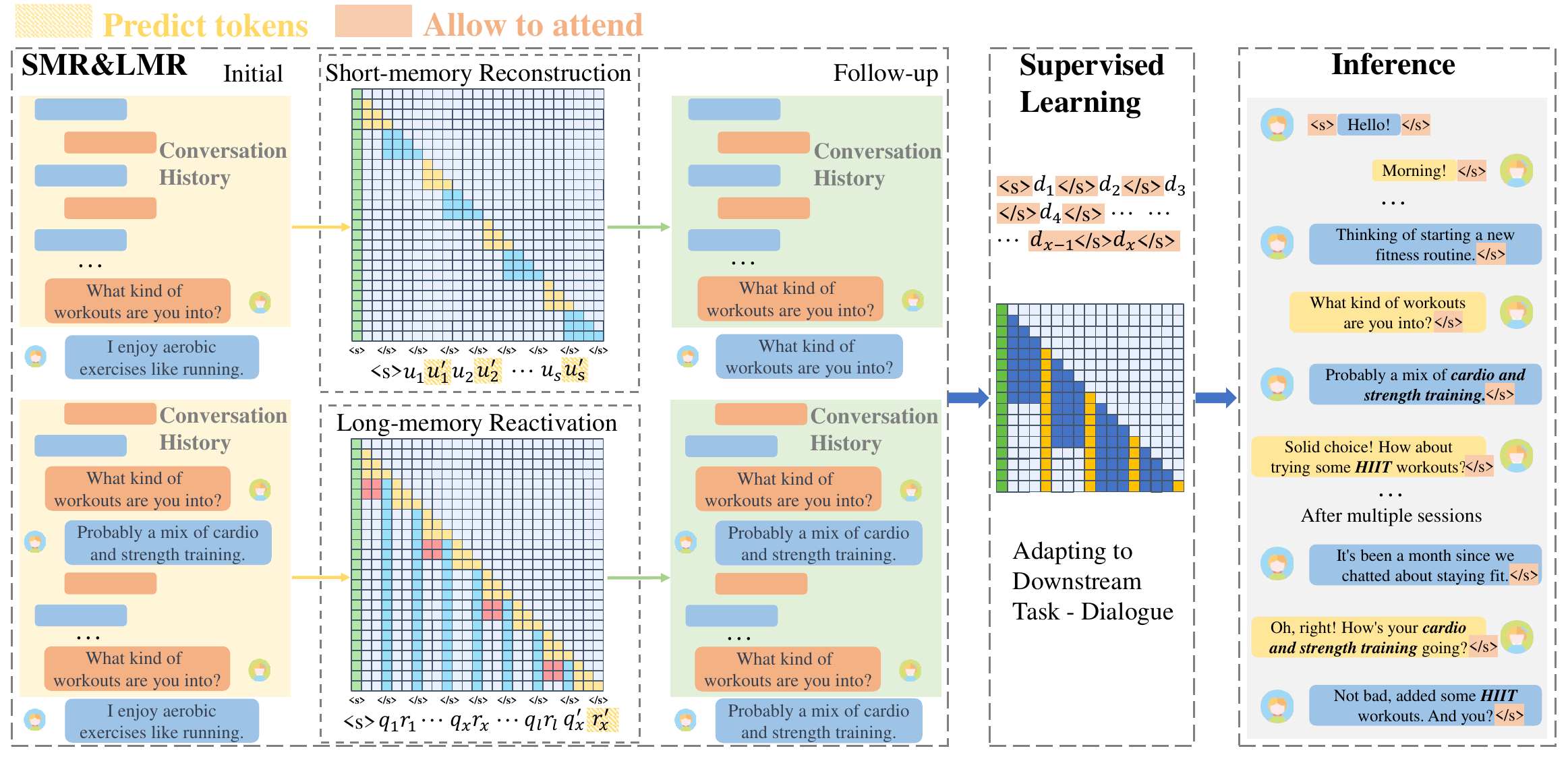}
  \caption{StreamingDialogue framework. SMR \& LMR strategies co-train the model by adjusting attention mechanisms. In supervised learning, the SMR \& LMR-trained model is fine-tuned with dialogue datasets. During inference, only specific tokens are cached, with critical historical dialogue information in bold italics for clarity.}
  \label{fig:framework}
\end{figure*}

\section{StreamingDialogue}

\subsection{Empirical observation}\label{Observation}

StreamingLLM~\cite{xiao2023efficient} focuses on initial tokens as attention sinks, i.e., initial tokens attract a significant amount of attention. After visualizing the attention maps of all layers and heads for both Llama-2-7B and Llama-2-7B-Chat, we observe a similar phenomenon in structured texts such as multi-turn dialogue, where LLMs tend to attend more to tokens used to separate dialogue when speakers switch (i.e., the end symbol ``</s>,'' newline symbol ``\textbackslash{}n,'' or other symbols, known as End-of-Utterance), and their attention aggregation is even greater than that of initial tokens (shown in Figure~\ref{fig:observe}). Based on the attention map, we suggest that each EoU captures the information of the current utterance, and EoUs are visible to subsequent utterances. These EoUs imply the potential for information aggregation, which we refer to as ``conversational attention sinks'' (conv-attn sinks).


According to the observation, rather than caching entire utterances to retain information as done in dense attention, we cache conv-attn sinks as a replacement. Let $T$ represent the number of utterances, and $L$ denote the average length of each utterance. By caching only the corresponding conv-attn sinks, the space complexity reduces from $O(TL)$ in dense attention to $O(T)$, and the time complexity from $O(T^2L^2)$ to $O(T^2L)$. Moreover, given the infrastructure of LLMs based long context modeling, our method is capable of efficiently handling dialogues with prolonged conversational histories. To this end, conv-attn sinks matter because they memorize the context information not only effectively, but also efficiently.

\subsection{Framework overview}

\subsubsection{Fine-tuning LLMs with conv-attn sinks}\label{tune-method}


We compress the content of each utterance into the subsequent conv-attn sink and recall historical information during the dialogue by attending to conv-attn sinks. To achieve this, we adjust the attention pattern \( A \in \{0, 1\}^{N \times N} \), where \( N \) represents the conversation length and 0 indicates masked attention values, i.e., specifying the specific keys and values that a query can attend to. Each token within an utterance can focus on the first token to uphold stable output during extended conversations, all preceding conv-attn sinks to extract historical context, and tokens from both the previous and current utterances to ensure continuity with the preceding utterance. Formally, we denote an utterance as \( u = d \)</s>, where \( d \) represents the dialogue content and </s> denotes the EoU token, a.k.a., conv-attn sink. Thus, a conversation can be organized as \( D = \) <s>\( u_1 u_2 ... u_t \), where \( t \) is the number of utterances. Attention mask matrix \( A \) is defined as:
\begin{equation*}
\label{base-attn}
    \begin{split}
        A_{i j} = \left\{\begin{aligned}
1, & \quad j = kl \leq i \ (k \in \mathbb{N}), \ 0 \leq i < N \\
1, & \quad 1 \leq j \leq i \leq l \\
1, & \quad j \neq kl \ (k \in \mathbb{N}), (\lceil \frac{i}{l} \rceil - 2) \cdot l < j \leq i < N \\
0,& \quad \text{otherwise},
\end{aligned}\right.
    \end{split}
\end{equation*}
where \( l \) denotes the average length of each utterance, $\mathbb{N}$ represents a non-negative integer, and $\lceil \cdot \rceil$ represents the ceiling function. During fine-tuning, all tokens are treated as predicted tokens to participate in the loss calculation.

While this method excels in managing long contexts compared to sparse attention methods like StreamingLLM, it falls short in characterizing short-term memories. To learn towards a more robust model with balanced memory capacity for both long and short memories, we propose two learning strategies to co-train the model and address these issues: short-memory reconstruction (SMR) and long-memory reactivation (LMR). The final version of StreamingDialogue is conducted in three stages, including SMR \& LMR, supervised learning, and inference, as illustrated in Figure \ref{fig:framework}.


\subsubsection{Short-memory reconstruction}
We propose a learning strategy to guide model behavior, enabling conv-attn sinks to consciously aggregate information through short-memory reconstruction (SMR). We reorganize data formats, modify the attention pattern, and adjust the loss function. More specifically, training samples are organized as \( D = \) <s>\( u_1 u_1' u_2 u_2' \ldots u_s u_s' \), where \(u_1, u_2, \ldots, u_s \) are randomly selected from the original dataset, and $s$ represents the number of randomly selected utterances. Each ``\( u u' \)'' pair can be regarded as a reconstruction task, where tokens in $u$ can attend to <s> and tokens that appear before the token in the current utterance. $u'$ can additionally attend to the conv-attn sink in $u$. The task objective is to reconstruct $u$ in $u'$, encouraging the conv-attn sink in $u$ to aggregate information from $u$ for utterance reconstruction. The attention pattern $A$ in SMR is defined as:
\begin{equation*}
    \begin{split}
        A_{i j} = \left\{\begin{aligned}
1, & \quad j = 0 \\
1, & \quad \lceil \frac{i}{l} \rceil = 2k \ (k \in \mathbb{N}^*), (\lceil \frac{i}{l} \rceil - 1) \cdot l \leq j \leq i < N \\
1, & \quad \lceil \frac{i}{l} \rceil = 2k + 1 \ (k \in \mathbb{N}), (\lceil \frac{i}{l} \rceil - 1) \cdot l < j \leq i < N \\
0,& \quad \text{otherwise},
\end{aligned}\right.
    \end{split}
\end{equation*}
where $\mathbb{N}^*$ represents a positive integer.

Since the goal is to reconstruct the contents of \(u_1', u_2', \ldots, u_s' \) into \(u_1, u_2, \ldots, u_s \), the loss calculation is defined as:
{
\begin{equation*}
    \mathcal{L}_{\text{SMR}}=-\sum_{(x, y) \in \mathcal{Z}} \sum_{t=1}^{|y|} \log \left(P_{\Phi}\left(y_{t} \mid x, y_{<t}\right)\right),
\end{equation*}
}

where $\mathcal{Z}=\left\{\left(x_{i}, y_{i}\right)\right\}_{i=1, \ldots, N}$ denotes the set of (\(u, u'\)) pairs. $x$ denotes the target utterance $u$, and $y$ represents the reconstructed utterance $u'$. The model learns to aggregate information into conv-attn sinks during SMR with minimal training.

\subsubsection{Long-memory reactivation}
The model is required to both aggregate dialogue information into conv-attn sinks and retrieve information from them. To ensure consistency in multi-turn dialogue, our proposed model must efficiently extract long context information during dialogue generation. Therefore, we introduce long-memory reactivation (LMR) to enhance its long-term memory capability.


Each pair of utterances in the dialogue dataset, denoted as $qr$, represents a query-response pair with $q$ and $r$ from distinct roles. We organize training samples as \(D = \) <s>\(q_1 r_1 \ldots q_x r_x \ldots q_l r_l q_x' r_x'\), with $l$ denoting the number of training pairs. Each pair \(q_x' r_x'\) at the end of the sample is randomly selected from historical dialogues. 

We design a response recall task where the goal is to recall \(r_x'\) from the historical context \(q_x r_x\) given query \(q_x'\). Concurrently, we adjust $A$ so that each utterance can only attend to the first token, all conv-attn sinks, and itself. Moreover, each response in a training pair can attend to the corresponding query, while the conv-attn sink of the response is restricted to attending only to the response itself, ensuring that the conv-attn sink aggregates information solely from its associated utterance, i.e.,
\begin{equation*}
    \begin{split}
        A_{i j} = \left\{\begin{aligned}
1, & \quad j = 0 \\
1, & \quad \lceil \frac{i}{l} \rceil = 2k \ (k \in \mathbb{N}^*), (\lceil \frac{i}{l} \rceil - 2) \cdot l < j \leq i < N \\
1, & \quad j = kl \ (k \in \mathbb{N}^*) \ \; \text{or} \; j > \lceil \frac{i}{l} \rceil - 1, \ j \leq i < N \\
0,& \quad \text{otherwise}.
\end{aligned}\right.
    \end{split}
\end{equation*}

Since the objective is to evoke historical dialogue within the response of the last training pair in the sample, the loss function of LMR is defined as:
{
\begin{equation*}
    \mathcal{L}_{LMR}=-\sum_{(m, n)} \sum_{t=1}^{|n|} \log \left(P_{\Phi}\left(n_{t} \mid m, n_{<t}\right)\right),
\end{equation*}
}where $m$ represents <s>\(q_1 r_1 \ldots q_x r_x \ldots q_l r_l q_x' \), and $n$ denotes \( r_x'\).

The model now effectively utilizes historical information through LMR. We co-train the model using SMR and LMR, fully harnessing the information aggregation potential of conv-attn sinks and enhancing both short and long-term memory capability of the proposed model.

Following SMR \& LMR, the model requires additional refinement through fine-tuning on dialogues using the methods outlined in Section \ref{tune-method}.


\begin{table}[!h]
\caption{Main results on the PersonaChat and MSC datasets. \( \downarrow \) indicates lower values are better, while \( \uparrow \) indicates the opposite. The best result for each metric is presented in bold, while the second-best one is underlined. * indicates significance ($p < 0.05$) via pairwise $t$-test compared to other methods. ``PC'' denotes PersonaChat and ``StrLLM'' represents StreamingLLM.}
\label{tab:Mainresults}
\centering
\resizebox{\linewidth}{!}{
\begin{tabular}{llcccccccccccc}
\toprule
\multirow{2}{*}{Data} & \multirow{2}{*}{Method} & PPL & \multicolumn{3}{c}{BLEU (\%)} & \multicolumn{3}{c}{ROUGE (\%)} & \multicolumn{3}{c}{Distinct (\%)} & USL-H (\%) & Dial-M\\
 &  & \( \downarrow \) & B-avg \( \uparrow \) & B-1 \( \uparrow \) & B-2 \( \uparrow \) & R-1 \( \uparrow \) & R-2 \( \uparrow \) & R-L \( \uparrow \) & D-1 \( \uparrow \) & D-2 \( \uparrow \) & D-3 \( \uparrow \) & \( \uparrow \) & \( \downarrow \)\\
\midrule
\multirow{6}{*}{PC} & Dense & ~~\textbf{8.41*} & 13.15 & ~~49.30* & ~~20.05* & \textbf{13.98} & \textbf{3.07} & \textbf{13.44} & ~~\textbf{16.37*}& ~~\textbf{41.61*} & ~~\textbf{63.36*} & ~~14.21* & ~~2.38*\\
\cmidrule{2-14}
 & Local & ~~11.59* & 13.01 & \underline{50.78} & ~~20.13* & 13.83 & 2.69 & 13.29 & ~~12.49* & ~~32.17* & ~~51.12* & ~~17.35* & \textbf{2.07}\\
 & Big Bird & ~~9.00* & ~~12.93* & ~~50.00* & ~~20.52* & 13.78 & 2.64 & 13.33 & ~~11.83* & ~~32.46* & ~~52.17* & ~~16.95* & ~~2.37*\\
 & StrLLM & 8.96 & \underline{13.16} & 50.15 & \underline{20.68} & 13.94 & 2.73 & 13.36 & ~~12.00* & ~~32.64* & ~~52.36* & ~~\underline{17.63*} & ~~2.30*\\
\cmidrule{2-14}
 & MemBART & ~~13.15* & ~~11.18* & ~~46.63* & ~~17.65* & 13.11 & 2.56 & 12.78 & ~~12.86* & ~~30.87* & ~~48.86* & ~~12.23* & ~~2.49*\\
 \cmidrule{2-14}
 & Ours  & \underline{8.71} & \textbf{13.63} & \textbf{51.27} & \textbf{20.77} & \underline{13.96} & \underline{3.05} & \underline{13.43} & \underline{14.43} & \underline{37.23} & \underline{58.07} & \textbf{17.96} & \underline{2.10}\\
 \midrule
\multirow{6}{*}{MSC} & Dense & \textbf{7.58} & \textbf{19.47} & \textbf{52.22} & \textbf{28.41} & \underline{16.93} & \textbf{2.92} & \underline{15.48} & ~~\textbf{12.85}* & ~~\textbf{37.75}* & ~~\textbf{57.51}* & ~~\underline{90.11}* & ~~1.94*\\
 \cmidrule{2-14}
 & Local & ~~8.92* & ~~13.34* & ~~41.14* & ~~20.44* & ~~13.48* & ~~1.88* & ~~12.61* & ~~7.89* & ~~22.71* & ~~35.89* & ~~76.68* & ~~2.15*\\
 & Big Bird & ~~8.42* & ~~16.54* & ~~46.63* & ~~24.77* & ~~15.32* & 2.34 & ~~14.15* & ~~8.72* & ~~25.81* & ~~40.34* & ~~85.30* & \underline{1.72}\\
 & StrLLM & ~~8.38* & ~~16.76* & ~~47.54* & ~~25.08* & ~~15.25* & ~~2.44* & ~~14.21* & ~~9.18* & ~~26.93* & ~~41.62* & ~~86.91* & \textbf{1.71}\\
 \cmidrule{2-14}
 & MemBART & ~~13.73* & ~~17.11* & ~~49.78* & ~~25.82* & ~~14.93* & 2.61 & ~~13.76* & ~~10.86* & ~~30.55* & ~~47.37* & ~~85.13* & ~~1.97* \\
 \cmidrule{2-14}
 & Ours & \underline{7.99} & \underline{19.33} & \underline{51.49} & \underline{28.12} & \textbf{17.18} & \underline{2.77} & \textbf{15.86} & \underline{11.54} & \underline{32.58} & \underline{50.27} & \textbf{90.48} & 1.76 \\
\bottomrule
\end{tabular}}
\end{table}

\section{Experiments}\label{Experiments}
\subsection{Experimental setup}
\paragraph{Datasets \& baselines}We conduct experiments on PersonaChat \citep{zhang-etal-2018-personalizing}, Multi-Session Chat (MSC) \citep{xu-etal-2022-beyond}, Topical-Chat \citep{gopalakrishnan2019topical} and MultiWOZ \citep{budzianowski2018multiwoz} datasets. MSC, known for its extended conversational context, differs from PersonaChat, which is single-session. MSC's training set includes up to 4 sessions, and the test set comprises 5 sessions. Please refer to Appendix \ref{Datasetdetails} for detailed information on the datasets. We use only the dialogue portions of all datasets and explore the impact of additional knowledge from various datasets in Appendix \ref{Additional_Knowledge}.

Baseline methods include dense attention \citep{vaswani2017attention} for capturing all information, local attention \citep{beltagy2020longformer} with a fixed window size restriction, Big Bird \citep{zaheer2020big} combining sliding window, global, and random attention, and StreamingLLM \citep{xiao2023efficient} attending to attention sinks alongside the recent fixed window. Additionally, we compare our method with several methods that support long contexts, i.e., MemBART \citep{wu2022stateful}, a memory-augmented Transformer encoder-decoder model, and HRED \citep{10.1145/2806416.2806493} and VHRED \citep{serban2017hierarchical}, which utilize hierarchical encoders capable of encoding conversations of arbitrary length\footnote{Appendix \ref{Implementation} provides attention features and implementation details for both the baselines and StreamingDialogue}.

\paragraph{Evaluation metrics}We evaluate model performance on the dialogue generation task using several metrics: BLEU (B-avg / B-1 / B-2, where B-avg is the average of BLEU-1 to BLEU-4 scores) \citep{papineni2002bleu}, ROUGE (R-1 / R-2 / R-L) \citep{lin-2004-rouge}, and Distinct (D-1 / D-2 / D-3) \citep{li-etal-2016-diversity}. Additionally, we utilize two reference-free metrics specifically designed for dialogue quality assessment: USL-H \citep{phy2020deconstruct} and Dial-M \citep{dey2023dial}. Perplexity (PPL) is also computed. Furthermore, we report the C scores \citep{madotto-etal-2019-personalizing} for various models on PersonaChat to assess the consistency of the generated dialogues.

\subsection{Main results}
Table \ref{tab:Mainresults} shows evaluation results for each method on the test sets (partial results related to baselines HRED and VHRED, metric C score, and datasets Topical-Chat and MultiWOZ, are available in Appendix \ref{Additional_results}). For generation, we use the last utterance from each test set episode as the target for the model to generate. We also calculate the overall PPL for the entire test sets.

Our method outperforms sparse attention and memory-augmented baselines, achieving higher scores in BLEU, ROUGE, Distinct, and USL-H, while maintaining lower PPL and Dial-M metrics. For example, in MSC, StreamingDialogue demonstrates significant improvements over the second-best baseline, StreamingLLM, with B-avg increasing from 16.76\% to 19.33\% and R-L rising from 14.21\% to 15.86\%. In PersonaChat, specifically in D-2, StreamingDialogue increases from 32.64\% to 37.23\% compared to StreamingLLM. The notable superiority of StreamingDialogue can be attributed to its focus on conv-attn sinks, which compress historical information into them and cache them to enhance long-term memory, unlike baselines that rely on local attention windows and cannot handle extended dialogues effectively.

Furthermore, StreamingDialogue exhibits comparable performance to dense attention, e.g., in PersonaChat, the difference in ROUGE is less than 0.02\%. It also outperforms dense attention in terms of R-1 and R-L in MSC and achieves better BLEU scores on PersonaChat. This validation highlights the more accurate information conveyance capabilities of text generated by our method.

\begin{figure}[h]

  \centering
  \includegraphics[width=0.95\linewidth]
    {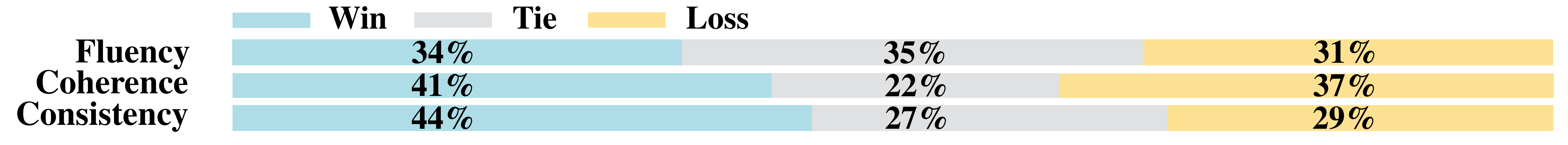}
  \caption{Fluency, coherence, and consistency in human evaluations: ours vs StreamingLLM.}
  \label{fig:human}
\end{figure}

\subsection{Human evaluation}
In human evaluation, we generate dialogues from 100 randomly selected episodes of the MSC test set. Four crowdsource evaluators compare our method with StreamingLLM in fluency, coherence, and consistency, categorizing the outcome as win, tie, or loss. Figure \ref{fig:human} demonstrates our method's superiority across all metrics, particularly in consistency, showcasing StreamingDialogue's superior long-term memory capacity. We apply Fleiss' kappa \cite{fleiss1971measuring} to measure the agreement among four annotators, yielding a result of 52.51\%. This indicates that the inter-annotator agreement is moderate ($\kappa \in [0.4, 0.6]$). More details on human evaluation are in Appendix \ref{humaneval}.

\begin{table}[h]
\caption{Ablation results on MSC with different learning strategies. ``Base'' denotes the model fine-tuned without SMR and LMR learning.}
\label{tab:ablation}
\centering
\small
\begin{tabular}{lcccc}
\toprule
Model & PPL & BLEU-avg & ROUGE-L & Distinct-3 \\
\midrule
Ours & 7.99 & 19.33 & 15.86 & 50.27 \\
\midrule
Base & 8.21 & 17.32 & 10.25 & 46.15 \\
LMR & 8.01 & 18.87 & 15.66 & 49.44 \\
SMR & 8.40 & 18.25 & 15.24 & 48.57 \\
\bottomrule
\end{tabular}
\end{table}

\subsection{Ablation results}


We conduct an experiment to test the effectiveness of SMR and LMR. Results, shown in Table \ref{tab:ablation}, highlight a significant decline in model performance when either strategy is ablated, indicating the importance of both strategies. The absence of SMR results in prominent declines in BLEU and ROUGE scores, indicating inadequate information aggregation in conv-attn sinks. Consequently, the model struggles to extract valuable information from conv-attn sinks during lengthy conversations, resulting in reduced text quality.

Similarly, without LMR, the model's performance declines significantly, indicating that relying solely on SMR leads to excessive guidance, limiting the model's ability in extended conversations. Thus, both SMR and LMR are crucial for enhancing information gathering and text extraction across conversations of long contexts.


\begin{figure}[htbp]
  \centering
  \begin{minipage}[t]{0.48\textwidth}
    \centering
    \includegraphics[width=\textwidth]{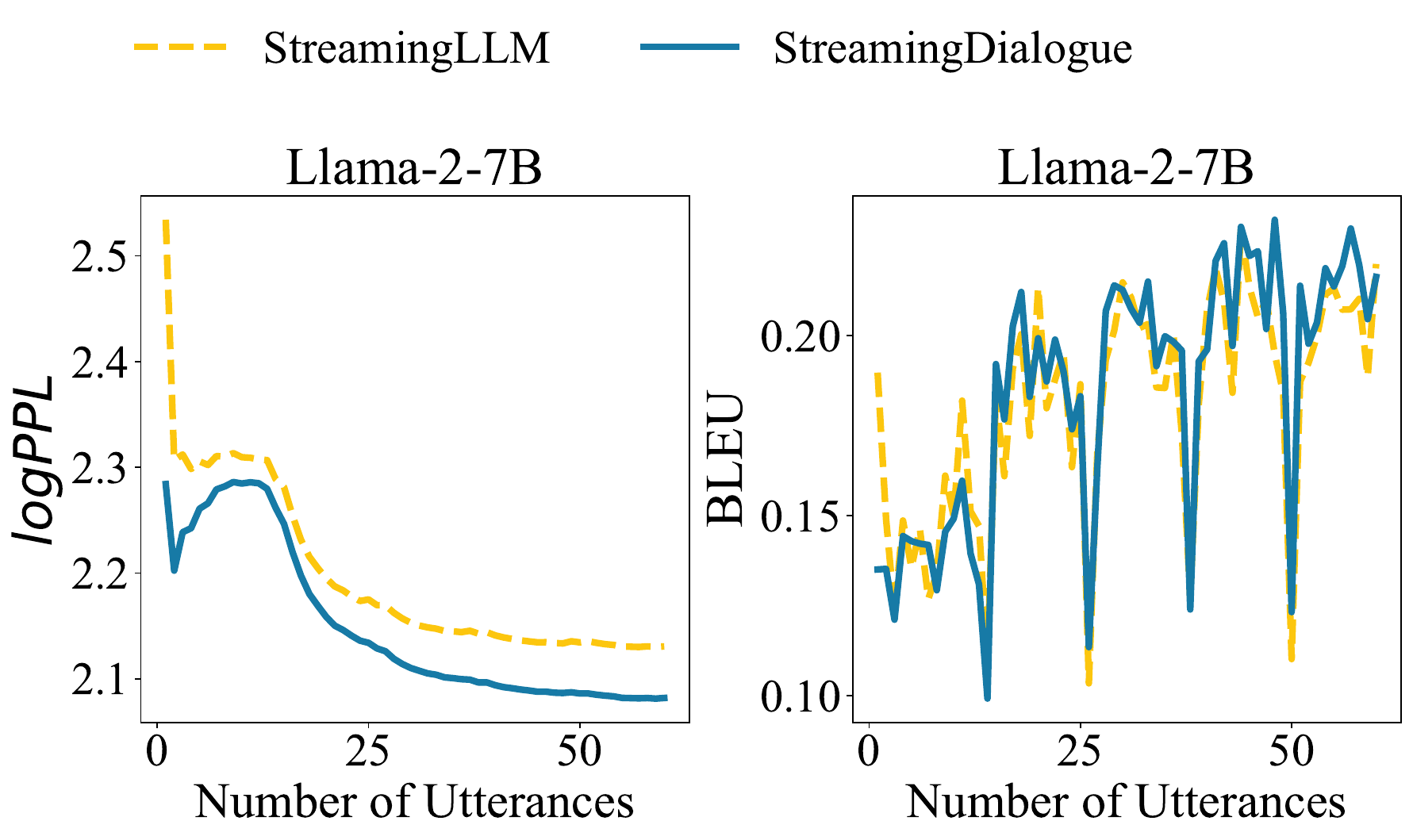}
    \caption{Average perplexity and BLEU for StreamingLLM and StreamingDialogue on the MSC test set across varying utterance counts.}
    \label{fig:ppl_bleu}
  \end{minipage}
  \hfill 
  \begin{minipage}[t]{0.48\textwidth}
    \centering
    \includegraphics[width=\textwidth]{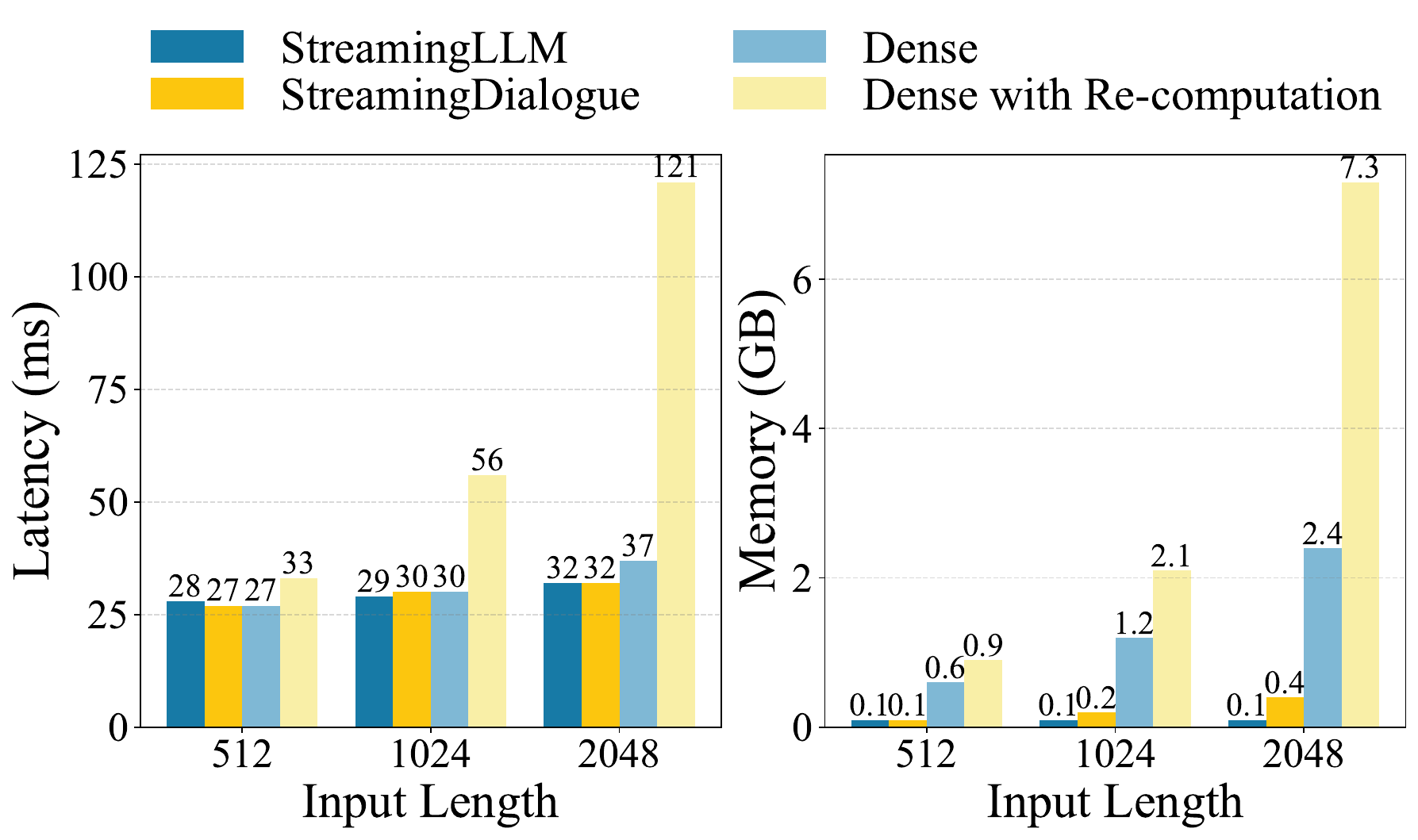}
    \caption{Per-token latency and memory usage by method on MSC for varying input lengths, with memory reported as total minus fixed.}
    \label{fig:speed}
  \end{minipage}
\end{figure}


\subsection{Performance on different context length}

\begin{figure}[!h]
  \centering
  \includegraphics[width=0.95\linewidth]{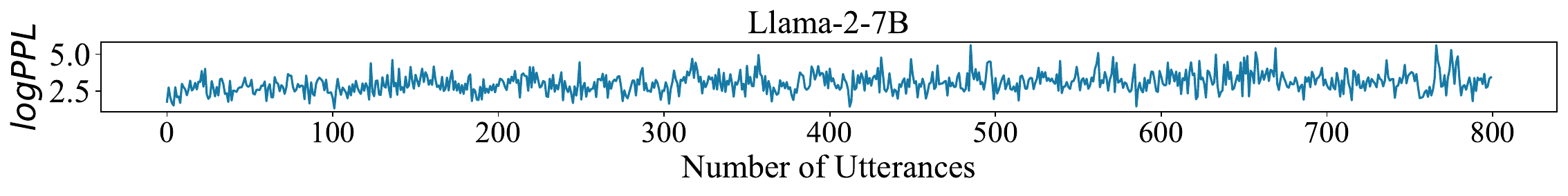}
  \caption{The perplexity for StreamingDialogue under the concatenated MSC test set, evaluating approximately 25K tokens.}
  \label{fig:ppl_long}
\end{figure}

We evaluate the model performance across different conversation lengths in terms of perplexity and BLEU under varying context length (i.e., the number of utterances in the dialogue context) during inference, as shown in Figure \ref{fig:ppl_bleu}. As dialogue length increases, StreamingDialogue exhibits greater superiority over StreamingLLM, with perplexity stabilizing and nearing convergence, and BLEU improving. Furthermore, StreamingDialogue maintains stable perplexity even with prolonged conversations over 25K tokens in inference (see Figure \ref{fig:ppl_long}). This highlights our method's stability in handling long dialogues and emphasizes the importance of conv-attn sinks in enhancing long-term memory.

\begin{table}[!h]
\caption{Results under the non-training setting on the MSC test set.}
\label{tab:non-training}
\centering
\resizebox{\linewidth}{!}{
\begin{tabular}{llcccccc}
\toprule
Model&Method & BLEU-avg & BLEU-1 & BLEU-2 & ROUGE-1 & ROUGE-2 & ROUGE-L\\
\midrule
\multirow{2}{*}{Llama-2-7B-Chat}&StreamingLLM & 20.16 & 51.18 & 29.99 & 15.90&1.92&14.26 \\
&Ours & 20.19 & 51.55 & 30.03 & 16.46 & 2.11&15.00 \\
\midrule
\multirow{2}{*}{Llama-3-8B-Instruct}&StreamingLLM & 16.48&39.68&24.63&16.88&1.93&15.47 \\
&Ours & 16.77&40.10&24.88&17.11&2.01&15.85 \\
\midrule
\multirow{2}{*}{Mistral-7B}&StreamingLLM & 12.75&42.86&19.99&12.58&1.83&11.73 \\
&Ours & 13.33&44.08&20.65&13.40&1.98&12.58\\
\bottomrule
\end{tabular}}
\end{table}
\subsection{Performance under the non-training setting}
We validate the performance of StreamingLLM and our method under a non-training setting on the MSC test set using Llama-2-7B-Chat, Llama-3-8B-Instruct \citep{dubey2024llama} and Mistral-7B \citep{jiang2023mistral}. As the table \ref{tab:non-training} shows, our StreamingDialogue, thanks to the conv-attn sinks, retains more complete historical information. Consequently, it consistently outperforms StreamingLLM. This demonstrates that applying conv-attn sinks for modeling long contexts remains effective even without any training.

\begin{table}[!h]
\caption{Dialogue reconstruction performance.}
\label{tab:reconstruct}
\centering
\small
\begin{tabular}{ccccc}
\toprule
BLEU-avg & BLEU-1 & BLEU-2 & ROUGE-1 & ROUGE-L \\
\midrule
68.02 & 89.19 & 76.83 & 76.79 & 72.94 \\
\bottomrule
\end{tabular}
\end{table}

\begin{figure}
  \centering
  \includegraphics[width=\linewidth]{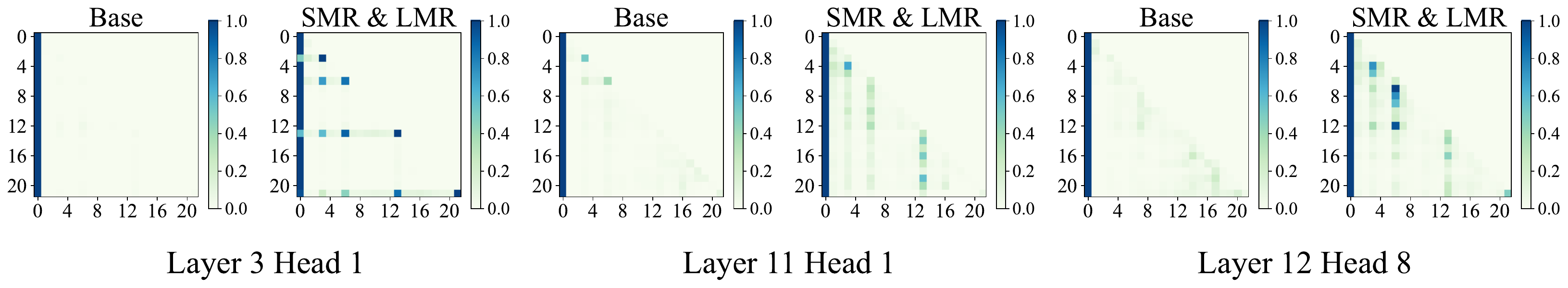}
  \caption{Comparison of attention maps before and after learning. ``Base'' denotes Llama-2-7B, while ``SMR \& LMR'' represents the model obtained post co-training with SMR and LMR on Llama-2-7B. The ``</s>'' positions in the encoded sentences are: 3, 6, 13, and 21.}
  \label{fig:6map}
\end{figure}

\begin{wrapfigure}[23]{r}{0.45\textwidth}
\vspace{-11pt}
  \centering
  \includegraphics[width=0.45\textwidth]{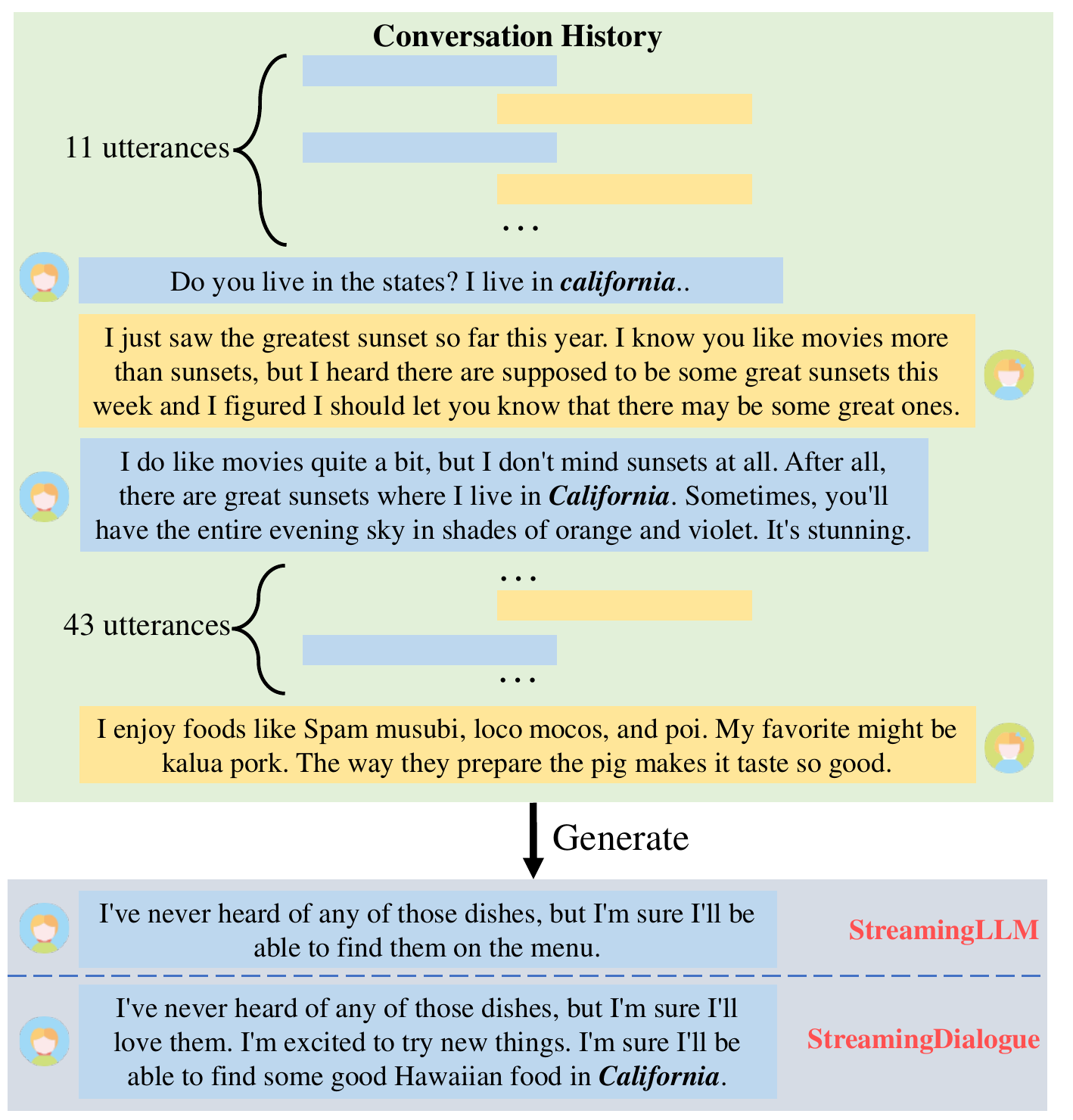}
  \caption{The generated dialogues by StreamingLLM and StreamingDialogue for the same input dialogue history from an MSC episode, with an average utterance length of $L=32$ tokens. Bold italic indicates key information in the dialogue.}
  \label{fig:case}
\end{wrapfigure}

\subsection{Information preservation}

To assess our method's information compression loss, we use SMR-trained models to reconstruct dialogue content from the MSC test set, leveraging only the conv-attn sink of each utterance. Randomly selecting 6,000 utterances from the MSC test set, we present the average results in Table \ref{tab:reconstruct}. Our method achieves a BLEU-1 score of 89.19\%, signifying effective compression of dialogue information with minimal losses.

\subsection{Speedup for inference}
Figure \ref{fig:speed} depicts the average per-token latency and memory usage during dialogue generation with NVIDIA A100 GPU using various methods. As input lengths increase, StreamingDialogue shows minimal growth in memory usage for caching conv-attn sinks, with latency exhibiting linear growth. This suggests that as dialogue length increases, StreamingDialogue's advantage becomes more promising. At an input length of 2,048, our method demonstrates a 6 $\times$ improvement in memory usage compared to dense attention and an 18 $\times$ improvement compared to dense attention with re-computation. In terms of per-token latency, our method shows a 4 $\times$ improvement compared to dense attention with re-computation. Moreover, our method maintains similar latency and memory usage as StreamingLLM as context length varies.

\subsection{Impacts of SMR \& LMR learning}

Since the motivation of SMR and LMR learning is to improve conv-attn sinks aggregation capability, we examine the attention maps after SMR and LMR co-training, comparing them with the base model to confirm enhancement.
Results are illustrated in Figure \ref{fig:6map}. Guided by SMR and LMR, the model's attention patterns transform into maps that sharply concentrate on conv-attn sinks, showcasing our effective enhancement of their information aggregation ability.

\subsection{Case study: memory capacity}
To validate StreamingDialogue's effectiveness in enhancing long-term memory, we conduct a case study comparing it with StreamingLLM. Figure \ref{fig:case} illustrates content generated by both methods. StreamingLLM responds solely based on recent utterances, lacking connection to distant context and coherence, thus reaffirming its unsuitability for open-domain dialogue. In contrast, StreamingDialogue effectively recalls distant historical information (e.g., 44 utterances ago), demonstrating the model's enhanced ability to remember long conversations through SMR and LMR.

\subsection{Generality of conv-attn sinks}\label{Generality}
To establish the generality of the conv-attn sink phenomenon, where separators attract more attention than other tokens within dialogues, we conduct both qualitative and quantitative analyses. Qualitatively, we visualize attention patterns across different training methods, attention mechanisms, and types of dataset constructions, demonstrating that this phenomenon persists regardless of these variables. Quantitative measures further support these findings, with a set threshold indicating significantly higher attention on separators than on other tokens. Detailed results, including visualizations and statistical data, are provided in Appendix \ref{ap-Generality}. This comprehensive analysis confirms the robustness of the conv-attn sink phenomenon across different settings and models.

\subsection{Analysis of EoU tokens' information aggregation capability}

To assess the effectiveness of EoU tokens in capturing dialogue information, we conduct experiments using an untrained Llama-2-7B-Chat model. The model focuses solely on EoU tokens and the last complete utterance. In a case study with the conversation: ``Did you have a caramel macchiato today?</s>Yes!</s>What kind of coffee did you have today?</s>,'' the model successfully identifies the key information, responding with, ``I had a delicious caramel macchiato this morning.''

We replicate this experiment across 10 different prompt formats, each containing 20 samples. The results indicate that 68\% of the model's responses accurately include essential information. Further details on these formats are available in Appendix \ref{prompt-format}.




\begin{wrapfigure}[12]{r}{0.45\textwidth}
\vspace{-15pt}
  \centering
  \includegraphics[width=0.45\textwidth]{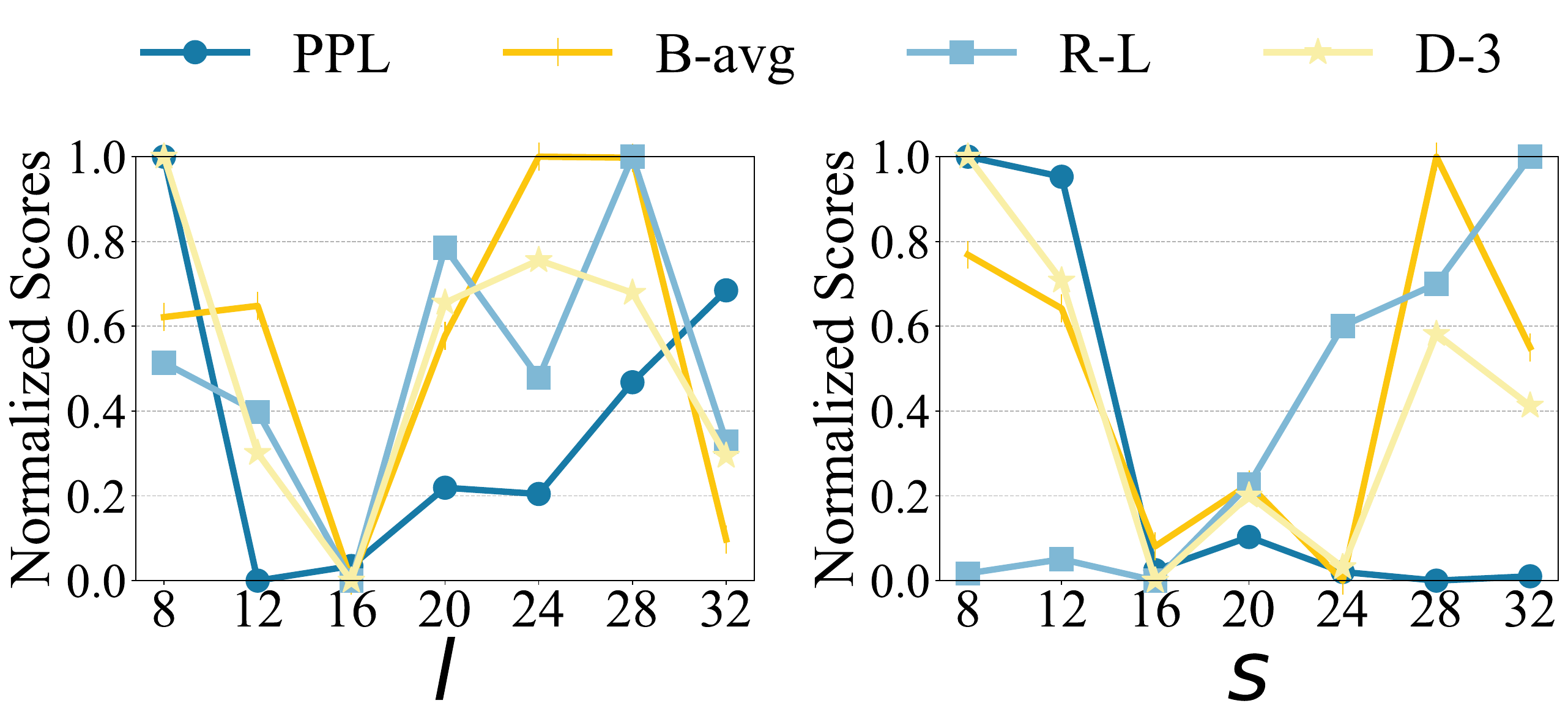}
  \caption{Normalized performance scores (PPL, B-avg, R-L, and D-3) on MSC for various $l$ with $s$ fixed at 28 and various $s$ with $l$ fixed at 24.}
  \label{fig:combinedSL}
\end{wrapfigure}

\subsection{Hyper-parameter sensitives}\label{Hyper-parameter}
We investigate the impact of two hyper-parameters in our method: the number of utterances in SMR samples ($s$) and the number of query-response pairs in LMR samples ($l$), both ranging from $\{8, 12, 16, 20, 24, 28, 32\}$. We maintain a constant total number of utterances in the training set, with examples like $s=8 \times \text{samples}=24,000$ and $s=12 \times \text{samples}=16,000$. Results shown in Figure \ref{fig:combinedSL} reveal that StreamingDialogue performs best with higher values of $s$ and $l$, optimally at $s \in \{28, 32\}$ and $l \in \{20, 24, 28\}$. This suggests that longer texts enhance the model’s learning.




\section{Conclusion}
Generating high-quality open-domain dialogues with prolonged contexts is quite challenging. Existing solutions, like dense attention, have efficiency issues. While StreamingLLM supports efficient language modeling, it struggles to preserve historical information, leading to low-quality generation in prolonged conversations. In this paper, we introduce StreamingDialogue, a framework capable of facilitating efficient and prolonged dialogue. By identifying separator tokens EoU as ``conv-attn sinks'' and compressing dialogue information into them with minimal losses, StreamingDialogue conserves memory, enhances efficiency, and augments long-term memory capabilities. Additionally, we propose two learning strategies to enhance conv-attn sink aggregation and memory reactivation. Our method shows better performance compared to strong baselines. In the future, we will explore extending StreamingDialogue towards never-ending dialogue in the context of lifelong learning.

\newpage

\begin{ack}
We sincerely thank Songhao Wu and Ang Lv for their valuable discussion and feedback on the manuscript.

This work was supported by the National Natural Science Foundation of China (NSFC Grant No. 62122089, U21B2027), Beijing Outstanding Young Scientist Program NO. BJJWZYJH012019100020098, and Intelligent Social Governance Platform, Major Innovation \& Planning Interdisciplinary Platform for the ``Double-First Class'' Initiative, Renmin University of China, the Fundamental Research Funds for the Central Universities, and the Research Funds of Renmin University of China.
\end{ack}

\bibliographystyle{unsrtnat.bst}
\bibliography{reference}


\newpage
\appendix
\section{Dataset details}
Refer to Table \ref{tab:dataset} for specific details regarding the PersonaChat, MSC, Topical-Chat and MultiWOZ datasets.
\label{Datasetdetails}
\begin{table}[h]
\caption{Details of dialogue datasets. We present the number of utterances (Utts.) and the average length per utterance (Avg. L) for each session in the training and test sets.}
\label{tab:dataset}
\centering
\small
\begin{tabular}{lccccccc}
\toprule
\multirow{2}{*}{Data} & \multirow{2}{*}{Data Type} & \multicolumn{2}{c}{Train} & \multicolumn{2}{c}{Test} \\
\multirow{2}{*}{} & \multirow{2}{*}{} & Utts. & Avg. L & Utts. & Avg. L \\
\midrule
PersonaChat & Total & 122499 & 13.59 & 14602 & 13.85 \\
\midrule
\multirow{6}{*}{MSC} & Session 1 & 59894 & 14.16 & 6572 & 15.47 \\
 & Session 2 & 46420 & 31.44 & 5939 & 30.86 \\
 & Session 3 & 47259 & 32.90 & 5924 & 32.94 \\
 & Session 4 & 11870 & 32.25 & 5940 & 34.67 \\
 & Session 5 & - & - & 5945 & 36.43 \\
 & Total & 165443 & 25.66 & 30320 & 29.77 \\
 \midrule
Topical-Chat & Total & 188378 & 26.76 & 11760 & 26.98 \\
\midrule
MultiWOZ & Total & 113552 & 18.92 & 14744 & 19.23 \\
\bottomrule
\end{tabular}
\end{table}

\section{Attention patterns \& implementation details}
\label{Implementation}
\begin{figure*}
 \centering
 \includegraphics[width=\linewidth]{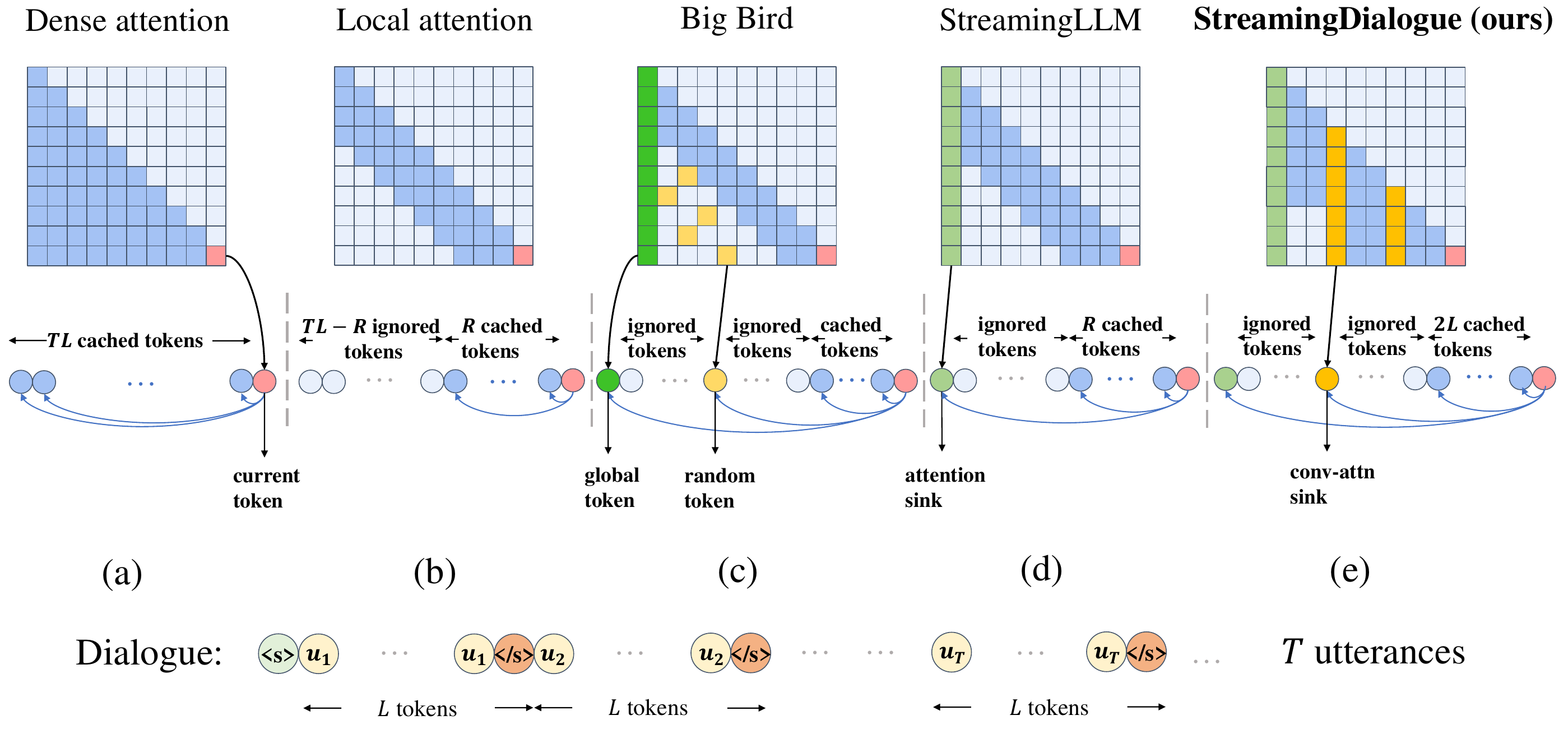}
 \caption{Attention maps' visualization of StreamingDialogue and various other methods. In a dialogue with $T$ utterances, each averaging $L$ tokens, dense attention caches $TL$ tokens, local attention caches $R$ tokens (where $R$ is the window size), Big Bird caches global size $+$ random size $+$ $R$ tokens, StreamingLLM caches $R+1$ tokens, and StreamingDialogue requires caching up to $1+T+2L$ tokens.}
 \label{fig:attn_all}
\end{figure*}
We visualize the attention maps of StreamingDialogue and several other baselines focusing on different attention patterns, as shown in Figure \ref{fig:attn_all}. Below are the details of the experiments for each method.

During the SMR \& LMR phase, we construct a short-memory reconstruction dataset comprising $n_s=6857$ samples, each containing a random selection of $s=28$ utterances from dialogue datasets. Simultaneously, for the long-memory reactivation dataset, we generate $n_l=8000$ samples, each consisting of a random selection of $l=24$ query-response pairs from the full set of query-response pairs in dialogue datasets, with one additional pair randomly chosen from the $l$ pairs appended to the end of each sample. The SMR and LMR datasets are merged and shuffled for co-training Llama-2-7B. We only train the attention layer for 1 epoch, with the learning rate set to 5e-5, utilizing cosine annealing for adjusting the learning rate, and setting the warm-up step to 0. The SMR \& LMR phase requires about 2 hours on two A100-40G GPUs.

In the supervised learning phase, StreamingDialogue undergoes fine-tuning based on the model trained with SMR \& LMR, while baselines focusing on different attention patterns are fine-tuned on Llama-2-7B. All models fine-tune only the attention layer for 2 epochs, with the learning rate set to 5e-5, utilizing cosine annealing for adjusting the learning rate, and setting the warm-up step to 0. This phase demands only approximately 1 hour on two A100-40G GPUs. MemBART, HRED and VHRED can be fine-tuned directly on dialogue datasets.

In inference, we set the maximum generation length to 120 and report the average results of all episodes. Dialogue generation takes only about 15 minutes on a single A100-40G GPU.



\section{Additional results for main evaluation}\label{Additional_results}
Table \ref{tab:Additional} presents the experimental results incorporating baselines HRED and VHRED. Our method achieves favorable results compared to any baseline across all metrics.
\begin{table}[!h]
\caption{Results on the MSC dataset. \( \downarrow \) indicates lower values are better, while \( \uparrow \) indicates the opposite. The highest-performing result for each metric is highlighted in bold, and the second-highest is underlined.}
\label{tab:Additional}
\centering
\small
\begin{tabular}{lccccccc}
\toprule
Method & B-avg \( \uparrow \) & R-1 \( \uparrow \) & R-2 \( \uparrow \) & D-1 \( \uparrow \) & D-2 \( \uparrow \) & USL-H \( \uparrow \) & Dial-M \( \downarrow \)\\
\midrule
StreamingLLM & 16.76 & \underline{15.25} & \underline{2.44} & \underline{9.18} & \underline{26.93} & \underline{86.91} & \textbf{1.71} \\
HRED & 15.72 & 14.75 & 1.85 & 7.37 & 20.91 & 58.70 & 2.13 \\
VHRED & \underline{17.02} & 15.16 & 1.48 & 5.28 & 14.72 & 59.31 & 2.35 \\
\midrule
Ours & \textbf{19.33} & \textbf{17.18} & \textbf{2.77} & \textbf{11.54} & \textbf{32.58} & \textbf{90.48} & \underline{1.76} \\
\bottomrule
\end{tabular}
\end{table}

Table \ref{tab:cscore} reports the results of the C score on PersonaChat. Our StreamingDialogue still achieves the best results among the baselines, except for dense attention. As an efficient algorithm, our method can significantly improve the speed compared to dense attention while maintaining the contextual and character consistency of long conversations.
\begin{table}[!h]
\caption{Results of the C score on the PersonaChat dataset. \( \uparrow \) indicates higher values are better.}
\label{tab:cscore}
\centering
\small
\begin{tabular}{lcccccc}
\toprule
Method & Dense & Local & Big Bird & StreamingLLM & MemBART & Ours \\
\midrule
C (\%) \( \uparrow \) & 3.10 & -3.40 & -4.00 & -4.70 & 0.77 & 2.70 \\
\bottomrule
\end{tabular}
\vspace{-10pt}
\end{table}

Table \ref{tab:MultiWOZ} presents the results for the Topical-Chat and MultiWOZ datasets, where our method outperforms all strong baselines by retaining more complete historical information.

\begin{table}[!h]
\caption{Results on the Topical-Chat and MultiWOZ datasets. \( \downarrow \) indicates lower values are better, while \( \uparrow \) indicates the opposite. The highest-performing result for each metric is highlighted in bold, and the second-highest is underlined.}
\label{tab:MultiWOZ}
\centering
\small
\begin{tabular}{llccccccc}
\toprule
Data & Method & PPL \( \downarrow \) & ROUGE-1 \( \uparrow \) & ROUGE-2 \( \uparrow \) & ROUGE-L \( \uparrow \) & Dial-M \( \downarrow \)\\
\midrule
\multirow{6}{*}{Topical-Chat} & Dense & \textbf{9.49}&\textbf{15.70}&\underline{3.65}&\textbf{14.88}&3.09\\
& Local	& 27.55	& 12.60	& 2.09 & 10.37&7.02 \\
& Big Bird&10.36&14.21&3.55&11.79&3.01 \\
&StreamingLLM&10.34&14.25&3.55&11.84&3.05 \\
& MemBART&12.54&13.86&2.98&13.18&\underline{2.83}\\
& Ours&\underline{9.80}&\underline{15.46}&\textbf{3.99}&\underline{14.37}&\textbf{2.66}\\
\midrule
\multirow{6}{*}{MultiWOZ} &Dense&\underline{4.51}&\underline{24.79}&\underline{13.93}&\underline{24.67}&\underline{2.27}\\
& Local&5.38&24.26&13.47&24.15&2.45\\
& Big Bird&4.79&24.38&13.26&24.30&2.51\\
& StreamingLLM&4.76&23.66&13.09&23.41&2.47\\
& MemBART&5.36&20.05&12.41&19.94&2.37\\
& Ours&\textbf{4.34}&\textbf{25.26}&\textbf{14.27}&\textbf{25.20}&\textbf{2.25}\\
\bottomrule
\end{tabular}
\end{table}

\section{Human evaluation details}
\label{humaneval}
\subsection{Information about evaluators}
We engage four crowdsource evaluators to assess the performance of our method. These evaluators exhibit outstanding English proficiency, enabling them to accurately discern subtle nuances and meanings in language expressions. Moreover, they possess a comprehensive understanding of the distinctions among fluency, coherence, and consistency, and are adept at determining which response is superior. In terms of human evaluation, we pass the review by relevant institutions, and we anonymize all evaluators' responses. After a reasonable assessment of the workload, we pay each evaluator \$1.70 per 10 samples.

\subsection{Task description}
Table \ref{tab:Task} presents the detailed task description provided to evaluators, where responses generated by different models are randomized in each sample.

\begin{table}[!h]
\caption{Task description provided to evaluators in human evaluation.}
\label{tab:Task}
\centering
\small
\begin{tabularx}{\textwidth}{X}  
\toprule
Task description \\
\midrule
We aim to evaluate the quality of dialogues generated by various models, specifically focusing on fluency, coherence, and consistency. You will be presented with a dialogue history followed by two responses generated by different models for the latest utterance. For each evaluation metric—fluency, coherence, and consistency—please identify the superior response. Assign a ``1'' if the first response is better, a ``2'' if the second response is better, and ``0'' if both responses are of equal quality. Ensure that your selections clearly reflect which response better meets each metric. \\
\\
The specific descriptions of the three metrics are as follows:\\
\\
(1) \textbf{Fluency} assesses whether the response itself is well-written and grammatically correct. \\
(2) \textbf{Coherence} refers to the response being relevant to the content of the historical dialogue. \\
(3) \textbf{Consistency} requires the response to remain consistent with the persona information and objective facts from the historical dialogue. \\
\bottomrule
\end{tabularx}
\end{table}

\begin{figure}[!h]
  \centering
  \includegraphics[width=0.95\linewidth]{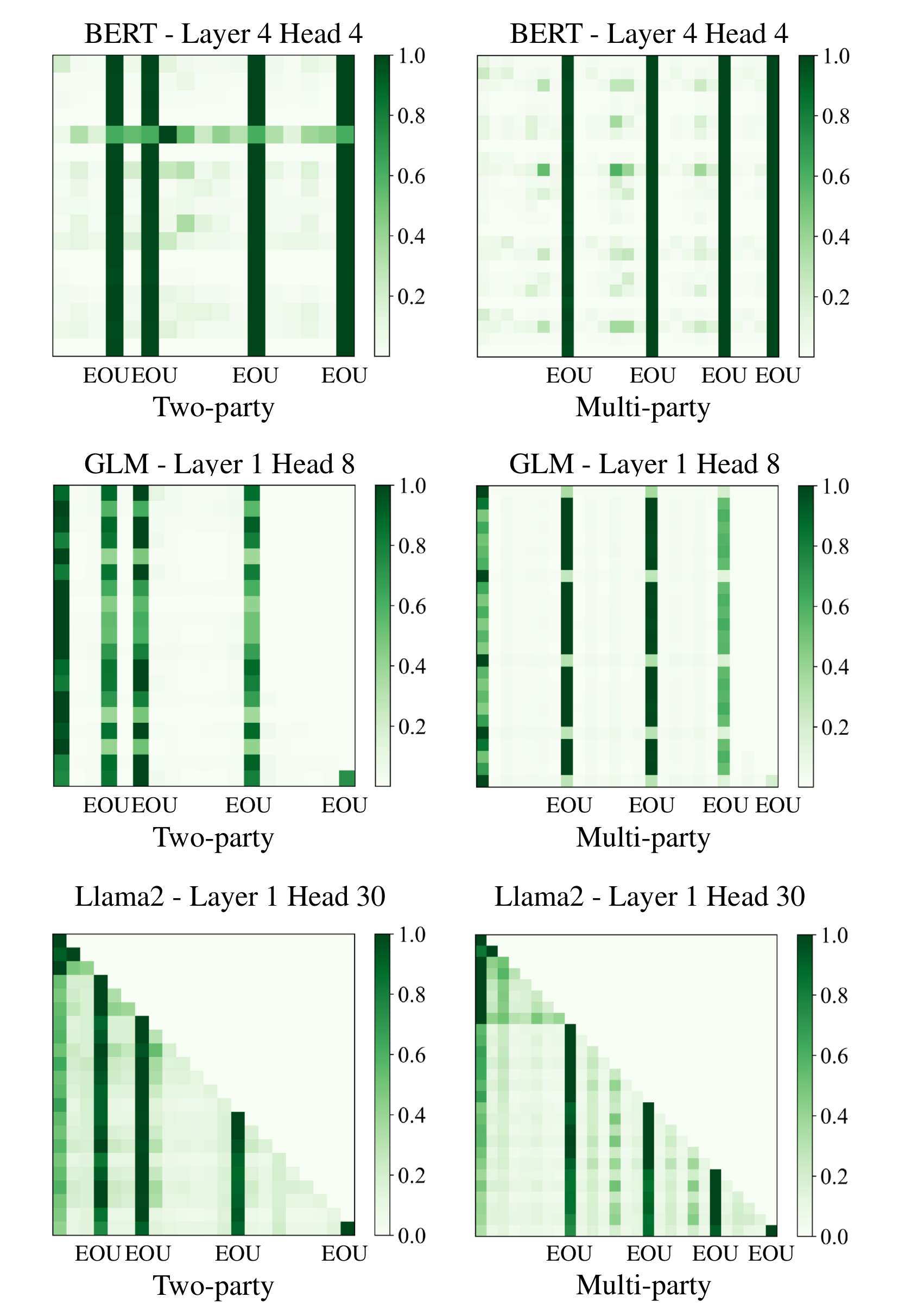}
  \caption{Attention maps under different settings.}
  \label{fig:multimodel}
\end{figure}

\section{Details on the generality of conv-attn sinks}
\label{ap-Generality}
To demonstrate the generality of the conv-attn sink phenomenon, i.e., the model aggregates more attention on separators than on other words and tokens, we provide both qualitative and quantitative analysis.

For \textbf{qualitative analysis}, we visualize the attentions using various training methods, attention mechanisms, and dataset constructions to observe the behavior of attention patterns.

(1) Regarding the independency on the training methods, we compare Llama2 with GLM \citep{du-etal-2022-glm}. Llama2 is pretrained by the next-token prediction objective, while GLM is pretrained on an autoregressive blank infilling objective.

(2) Regarding the independency on the attention mechanisms, we compare Llama2 with BERT \citep{devlin-etal-2019-bert}. Llama2 uses a unidirectional causal attention mechanism, while BERT uses a bidirectional attention mechanism.

(3) Reagarding the independency on the dataset constructions, we compare two-person dialogues with multi-party dialogues. Otherwise, we also have validated that the different separators for orgnizing the dialogue also exhibit the same phenomenon, as illustrated in Figure \ref{fig:observe}.

The visualization results under different settings are shown in Figure \ref{fig:multimodel}, which demonstrates the consistent conclusion that separators will attract more attention than other tokens in the dialogue. For the two-person dialogue, the data we use is ``<BOS>Hey!<EOS>Morning<EOS>Wanna grab coffee later?<EOS>Totally! Starbucks?<EOS>,'' and for the multi-party dialogue, the data we use is ``<BOS>Alice: How was your weekend<EOS>Bob: Good, yours?<EOS>Charlie: Great, thanks<EOS>Alice: Awesome<EOS>.'' ``<BOS>'' denotes the start symbol, and ``<EOS>'' denotes the end symbol.

\begin{table}[!h]
\caption{Proportion of attention heads exhibiting the conv-attn sink phenomenon across models.}
\label{tab:multimodel}
\centering
\small
\begin{tabular}{lccc}
\toprule
Model & Llama2 & GLM & BERT \\
\midrule
Proportion (\%) & 14.36 & 16.32 & 79.86 \\
\bottomrule
\end{tabular}
\end{table}

For \textbf{quantitative analysis}, we set a threshold indicating the occurrence of the conv-attn sink phenomenon in an attention head. This threshold is defined as the average attention aggregated by conv-attn sinks being three times or more than that aggregated by normal tokens. We report the proportion of these conv-attn heads among all attention heads using three heterogeneous models, as depicted in the Table \ref{tab:multimodel}. This illustrates the universality of the conv-attn sink. Furthermore, we observe a stronger tendency of the conv-attn sink in the BERT model, warranting further investigation in our future work.

\section{Impact of additional knowledge}\label{Additional_Knowledge}
The original datasets used for both training and testing contain some additional knowledge: Persona-Chat includes personas, Topical-Chat includes grounding knowledge, and MultiWOZ includes belief states. Our main experiments utilize only the dialogue portions of these datasets, without leveraging the additional knowledge. To make our results more compelling, we explore the impact of incorporating this extra knowledge into our experiments.

For MultiWOZ, we add the belief states before each corresponding utterance. The results are shown in the Table \ref{tab:MultiWOZ-extra}.

\begin{table}[!h]
\caption{The results of integrating the belief states on the MultiWOZ dataset.}
\label{tab:MultiWOZ-extra}
\centering
\small
\begin{tabular}{lccccccc}
\toprule
Method&PPL&BLEU-avg&BLEU-1&BLEU-2&Distinct-1&Distinct-2&Distinct-3 \\
\midrule
Dense&1.92&25.56&48.33&29.14&3.74&6.86&8.89\\
StreamingLLM&2.19&25.70&47.53&29.21&4.48&9.09&12.60\\
Ours&1.98&25.77&48.58&29.38&5.30&10.03&13.60\\
\bottomrule
\end{tabular}
\end{table}

Since our method retains historical information by compressing each utterance's information into conv-attn sinks, only the conv-attn sinks from the previous utterances will be attended to in subsequent utterances. Therefore, for Topical-Chat and Persona-Chat, we consider two settings:

(1) We treat each sentence of the grounding knowledge/persona profiles as an utterance, and the subsequent utterances can only attend to their conv-attn sinks. The results are shown in the Table \ref{tab:persona-extra1}.

\begin{table}[!h]
\caption{Results on the Topical-Chat and Persona-Chat datasets under the setting of treating each sentence of the grounding knowledge/persona profiles as an utterance.}
\label{tab:persona-extra1}
\centering
\small
\begin{tabular}{llcccc}
\toprule
Data&Method&PPL&Distinct-2&Distinct-3&Dial-M \\
\midrule
\multirow{3}{*}{PersonaChat}&Dense&7.19&43.56&66.27&2.53\\
&StreamingLLM&8.36&33.17&53.58&2.47\\
&Ours&7.60&39.16&61.06&2.36\\
\midrule
\multirow{3}{*}{Topical-Chat}&Dense&3.24&39.07&57.64&4.32\\
&StreamingLLM&8.31&16.87&23.56&3.72\\
&Ours&3.20&31.47&49.10&2.57\\
\bottomrule
\end{tabular}
\end{table}

(2) We use the grounding knowledge/persona profiles as a prompt: ``The conversation will be based on the following knowledge: <knowledge> {detailed knowledge} <conversation>'' in Topical-Chat and ``The conversation will be based on the following persona profile: <persona> {detailed persona profiles} <conversation>'' in Persona-Chat, allowing the subsequent utterances to fully attend to it. The results are shown in the Table \ref{tab:persona-extra2}.

\begin{table}[!h]
\caption{Results on the Topical-Chat and Persona-Chat datasets under the setting of treating the grounding knowledge/persona profiles as a prompt.}
\label{tab:persona-extra2}
\centering
\small
\begin{tabular}{llcccc}
\toprule
Data&Method&PPL&Distinct-2&Distinct-3&Dial-M \\
\midrule
\multirow{3}{*}{PersonaChat}&Dense&7.93&44.26&66.63&2.48\\
&StreamingLLM&7.99&36.40&57.44&2.91\\
&Ours&7.67&37.82&58.93&2.57\\
\midrule
\multirow{3}{*}{Topical-Chat}&Dense&11.64&36.98&54.96&4.60\\
&StreamingLLM&30.37&26.07&34.26&3.61\\
&Ours&10.21&32.16&50.41&2.97\\
\bottomrule
\end{tabular}
\end{table}

In a setting that includes grounding knowledge, our method consistently retains the memory of both grounding knowledge and historical dialogue. As a result, our method still outperforms the baseline, except for dense attention. As an efficient algorithm, our method can significantly improve speed compared to dense attention while maintaining the contextual and character consistency of long conversations.

\section{Examples of prompt formats}\label{prompt-format}

Examples of prompt formats are as follows, where the ``keywords'' will be replaced with specific content.

\begin{enumerate}
    \item ``template'': ``A and B went to PLACE today.</s>They had a great time.</s>Who did A go to PLACE with today?</s>'', \\
    ``keywords'': {``A'': ``person'', ``B'': ``person'', ``PLACE'': ``place''},\\
    ``answer key'': ``B''
    \item ``template'': ``B made A's favorite food, FOOD, today.</s>A was delighted.</s>What food did B make for A today?</s>'', \\
    ``keywords'': {``A'': ``person'', ``B'': ``person'', ``FOOD'': ``food''},\\
    ``answer key'': ``FOOD''
    \item ``template'': ``A was doing ACTIVITY when B called.</s>A had to stop and answer the call.</s>What was A doing when B called?</s>'', \\
    ``keywords'': {``A'': ``person'', ``B'': ``person'', ``ACTIVITY'': ``activity''},\\
    ``answer key'': ``ACTIVITY''
    \item ``template'': ``A bought a new ITEM today.</s>B was impressed by A's purchase.</s>What item did A buy today?</s>'', \\
    ``keywords'': {``A'': ``person'', ``B'': ``person'', ``ITEM'': ``item''},\\
    ``answer key'': ``ITEM''
    \item ``template'': ``A participated in an EVENT today.</s>B cheered them on.</s>What event did A participate in?</s>'', \\
    ``keywords'': {``A'': ``person'', ``B'': ``person'', ``EVENT'': ``event''},\\
    ``answer key'': ``EVENT''
\end{enumerate}

\section{Limitations}
\label{Limitations}
StreamingDialogue significantly reduces space and time complexity during the inference stage. Additionally, we can outperform the baseline under the non-training setting without additional cost. To optimize LLMs for the conv-attn sinks mode, we implement two learning strategies: short-memory reconstruction and long-memory reactivation. Consequently, this inevitably increases computational costs under the training setting, with the SMR and LMR phases requiring about two hours on two A100-40G GPUs.

While StreamingDialogue effectively enables prolonged conversations with long-term memory, there's merit in exploring selective caching of conv-attn sinks, focusing only on those aggregating key information. This will further enhance inference speed and reduce memory usage. Additionally, our utilization of dialogue structure is somewhat limited, and we aim to leverage conv-attn sinks to explore more intricate dialogue features in the future. Furthermore, evaluating our method across a wider range of structured texts will offer a more comprehensive assessment.

\section{Broader impacts and safety issues}\label{safety}
This paper identifies the conversational attention sink (conv-attn sink) phenomenon and proposes two learning strategies to naturally compress extended dialogue history into conv-attn sinks and effectively retrieve memories in subsequent conversations, thereby enabling prolonged streaming dialogue. Additionally, the datasets utilized, including PersonaChat and MSC, as well as the pre-trained models Llama-2-7B and Llama-2-7B-Chat, were sourced from their respective publishers through official open-source channels. Our enhancements are purely architectural and we will not release any new models or datasets. Nonetheless, the capability to efficiently compress and retrieve long dialogues may allow this technology to be used for monitoring conversations over extended periods, thus raising privacy concerns and the potential for surveillance without proper consent or transparency.

\newpage
\section*{NeurIPS Paper Checklist}

\begin{enumerate}

\item {\bf Claims}
    \item[] Question: Do the main claims made in the abstract and introduction accurately reflect the paper's contributions and scope?
    \item[] Answer: \answerYes{} 
    \item[] Justification: In the abstract, we claim that StreamingDialogue efficiently compresses dialogue history into conversational attention sinks with minimal losses, enhancing the model's long-term memory and facilitating prolonged streaming conversations.

\item {\bf Limitations}
    \item[] Question: Does the paper discuss the limitations of the work performed by the authors?
    \item[] Answer: \answerYes{} 
    \item[] Justification: We create a separate "Limitations" section, i.e., Section \ref{Limitations}.

\item {\bf Theory Assumptions and Proofs}
    \item[] Question: For each theoretical result, does the paper provide the full set of assumptions and a complete (and correct) proof?
    \item[] Answer: \answerNA{} 
    \item[] Justification: Our paper does not introduce theoretical results.

    \item {\bf Experimental Result Reproducibility}
    \item[] Question: Does the paper fully disclose all the information needed to reproduce the main experimental results of the paper to the extent that it affects the main claims and/or conclusions of the paper (regardless of whether the code and data are provided or not)?
    \item[] Answer: \answerYes{} 
    \item[] Justification: We have shared a URL that contains executable code, enabling easy reproduction of our experiments.

\item {\bf Open access to data and code}
    \item[] Question: Does the paper provide open access to the data and code, with sufficient instructions to faithfully reproduce the main experimental results, as described in supplemental material?
    \item[] Answer: \answerYes{} 
    \item[] Justification: We have open-sourced our code and shared the link.

\item {\bf Experimental Setting/Details}
    \item[] Question: Does the paper specify all the training and test details (e.g., data splits, hyperparameters, how they were chosen, type of optimizer, etc.) necessary to understand the results?
    \item[] Answer: \answerYes{} 
    \item[] Justification: See Section \ref{Experiments} and appendices.

\item {\bf Experiment Statistical Significance}
    \item[] Question: Does the paper report error bars suitably and correctly defined or other appropriate information about the statistical significance of the experiments?
    \item[] Answer: \answerYes{} 
    \item[] Justification: We present the statistical significance of the experiments in Table \ref{tab:Mainresults}.

\item {\bf Experiments Compute Resources}
    \item[] Question: For each experiment, does the paper provide sufficient information on the computer resources (type of compute workers, memory, time of execution) needed to reproduce the experiments?
    \item[] Answer: \answerYes{} 
    \item[] Justification: We provide detailed information on the computing resources required to reproduce the experiments in Section \ref{Implementation}.
    
\item {\bf Code Of Ethics}
    \item[] Question: Does the research conducted in the paper conform, in every respect, with the NeurIPS Code of Ethics \url{https://neurips.cc/public/EthicsGuidelines}?
    \item[] Answer: \answerYes{} 
    \item[] Justification: We have conformed the NeurIPS Code of Ethics.

\item {\bf Broader Impacts}
    \item[] Question: Does the paper discuss both potential positive societal impacts and negative societal impacts of the work performed?
    \item[] Answer: \answerYes{} 
    \item[] Justification: See Section \ref{safety}.
    
\item {\bf Safeguards}
    \item[] Question: Does the paper describe safeguards that have been put in place for responsible release of data or models that have a high risk for misuse (e.g., pretrained language models, image generators, or scraped datasets)?
    \item[] Answer: \answerYes{} 
    \item[] Justification: See Section \ref{safety}.

\item {\bf Licenses for existing assets}
    \item[] Question: Are the creators or original owners of assets (e.g., code, data, models), used in the paper, properly credited and are the license and terms of use explicitly mentioned and properly respected?
    \item[] Answer: \answerYes{} 
    \item[] Justification: We have correctly cited all the data, scripts, and models we used.

\item {\bf New Assets}
    \item[] Question: Are new assets introduced in the paper well documented and is the documentation provided alongside the assets?
    \item[] Answer: \answerYes{} 
    \item[] Justification: We have included a README document with our code.

\item {\bf Crowdsourcing and Research with Human Subjects}
    \item[] Question: For crowdsourcing experiments and research with human subjects, does the paper include the full text of instructions given to participants and screenshots, if applicable, as well as details about compensation (if any)? 
    \item[] Answer: \answerYes{} 
    \item[] Justification: See Section \ref{humaneval}.

\item {\bf Institutional Review Board (IRB) Approvals or Equivalent for Research with Human Subjects}
    \item[] Question: Does the paper describe potential risks incurred by study participants, whether such risks were disclosed to the subjects, and whether Institutional Review Board (IRB) approvals (or an equivalent approval/review based on the requirements of your country or institution) were obtained?
    \item[] Answer: \answerYes{} 
    \item[] Justification: See Section \ref{humaneval}.

\end{enumerate}

\end{document}